\documentclass[journal,twoside,web]{ieeecolor}
\usepackage{tmi}
\usepackage{cite}
\usepackage{amsmath,amssymb,amsfonts}
\usepackage{algorithmic}
\usepackage{graphicx}
\usepackage{textcomp}

\usepackage{hyperref}
\usepackage{algorithm}
\usepackage{booktabs}
\usepackage{multirow}

\def\BibTeX{{\rm B\kern-.05em{\sc i\kern-.025em b}\kern-.08em
    T\kern-.1667em\lower.7ex\hbox{E}\kern-.125emX}}
\begin{document}
\title{A Dual-branch Self-supervised Representation Learning Framework for Tumour Segmentation in Whole Slide Images}
\author{Hao Wang, \IEEEmembership{Student Member, IEEE}, Euijoon Ahn, and Jinman Kim, \IEEEmembership{Member, IEEE}
\thanks{This work was supported in part by the Australian Research Council (ARC).
(Corresponding author: Hao Wang and Jinman Kim.) }
\thanks{Hao Wang and Jinman Kim are with the Faculty of Engineering, School of Computer Science, The University of Sydney, Sydney, NSW 2006, Australia (e-mail: hwan7885@uni.sydney.edu.au; jinman.kim@sydney.edu.au).}
\thanks{Euijoon Ahn is with College of Science and Engineering, James Cook University, Cairns, QLD 4870, Australia (e-mail: euijoon.ahn@jcu.edu.au).}
}

\maketitle

\begin{abstract}
Supervised deep learning methods have achieved considerable success in medical image analysis, owing to the availability of large-scale and well-annotated datasets. However, creating such datasets for whole slide images (WSIs) in histopathology is a challenging task due to their gigapixel size. In recent years, self-supervised learning (SSL) has emerged as an alternative solution to reduce the annotation overheads in WSIs, as it does not require labels for training. These SSL approaches, however, are not designed for handling multi-resolution WSIs, which limits their performance in learning discriminative image features. In this paper, we propose a Dual-branch SSL Framework for WSI tumour segmentation (DSF-WSI) that can effectively learn image features from multi-resolution WSIs. Our DSF-WSI connected two branches and jointly learnt low and high resolution WSIs in a self-supervised manner. Moreover, we introduced a novel Context-Target Fusion Module (CTFM) and a masked jigsaw pretext task to align the learnt multi-resolution features. Furthermore, we designed a Dense SimSiam Learning (DSL) strategy to maximise the similarity of different views of WSIs, enabling the learnt representations to be more efficient and discriminative. We evaluated our method using two public datasets on breast and liver cancer segmentation tasks. The experiment results demonstrated that our DSF-WSI can effectively extract robust and efficient representations, which we validated through subsequent fine-tuning and semi-supervised settings. Our proposed method achieved better accuracy than other state-of-the-art approaches. Code is available at \url{https://github.com/Dylan-H-Wang/dsf-wsi}.
\end{abstract}

\begin{IEEEkeywords}
Histopathology, self-supervised representation learning, semantic segmentation, whole slide images.
\end{IEEEkeywords}

\section{Introduction}

\IEEEPARstart{W}{hole} slide images (WSI), also known as virtual microscopy, are a high-resolution digital image type produced by a complete microscope slide. They supply various microscopic views including nuclear atypia, degree of gland formation, mitosis and inflammation under different image resolutions, providing a thorough set of statistics about tissues and tumours. Pathologists use this information to assist with primary and secondary (consultation) diagnoses in pathology \cite{ref1}. In a standard WSI analysis, pathologists typically need to combine observations from multi-resolution WSIs due to the variety of tumour growth patterns. For example, WSIs with low resolution, which we refer to as the \textit{context images} in this paper, provide coarse-grained locations of tumours and the global architectural composition of tissue samples, such as duct presence. Pathologists then use high-resolution images of each region of interest (ROI) of the tumour part, which we refer to as the \textit{target images} in this paper, to analyse more specific information about cells, such as local cellular composition. However, manual analysis of WSIs is immensely time-consuming and laborious, requiring careful expert examinations \cite{ref38}. As a result, there has been sustained interest in recent years in building an automated computer-aided diagnosis (CAD) system for WSI. A core task in the WSI analysis is semantic segmentation of ROIs (e.g., tumours) that requires classification of each pixel of the WSI images. Accurate segmentation is important in histopathology for disease characterisation and diagnosis, assisting pathologists in making a final diagnosis with higher accuracy and less effort.

However, automatic and accurate segmentation is challenging for conventional machine learning methods due to the variations in cell size, shape, fuzzy boundaries, different cell colours and increasing input image resolutions. Over the last decades, deep learning techniques, such as Convolutional Neural Networks (CNNs) \cite{ref2}, have shown promising performance in improving WSI analysis. For example, Zhao et al. \cite{ref41} proposed a fully convolution network (FCN) utilising feature maps that have pyramid structures in each layer for efficient breast cancer segmentation in WSI. Similarly, Liu et al. \cite{ref42} designed a method that breaks whole images into small-sized image patches for fine-grained segmentation. Instead of feeding low-resolution whole images into the network, patch-based methods are capable of generating delicate segmentation maps by using high-resolution patches. Recently, Zhang et al. \cite{ref32} devised a dual-task which aims to solve detection (i.e., bounding boxes showing tumour locations) and segmentation tasks simultaneously to better learn image feature representation in WSI. Moreover, researchers also simulated the pathologists examination procedure and utilised multi-resolution WSIs to improve CNN performance \cite{ref12,ref13, ref34}. They proposed to apply multi-branch architecture to process different resolution images and showed the advantages of multi-resolution over single-resolution WSIs. While these approaches have shown good results, their performance is dependent on the availability of large-scale labelled data for model training. The annotation overheads, however, can be more severe for deep learning methods where experts need to attain accurate labels, and thus making it intensively burdensome to obtain high-quality large-scale datasets.

To alleviate the lack of data annotation, self-supervised learning (SSL), as a label-free algorithm, has received increasing attentions lately. SSL leverages a variety of data generated labels that are free of human annotations, e.g., predicting the rotation of randomly rotated images. Recent advancements of SSL \cite{ref5,ref6,ref7,ref8}, have successfully shown that CNNs are capable of learning meaningful image features without the need of manual labels, and the learnt representations by CNNs are shown to be effective and robust to carry transferable semantic or structural information in various image analysis tasks. Many recent SSL approaches \cite{ref43,ref44,ref45} also have been used to pre-train a model by leveraging all available unlabelled data for model initialisation. This initialisation is helpful to improve prediction performance when the model is later fine-tuned with smaller set of labelled data. Some preliminary works \cite{ref9,ref10,ref11} have shown promising outcomes by integrating SSL-pre-trained feature extractor to mitigate data scarcity and domain shift (i.e., transferring from natural images to WSIs) problem in histopathology. For example, Ciga et al. \cite{ref9} pre-trained a model based on the SSL algorithm SimCLR \cite{ref7} with a large hybrid WSI dataset and achieved a better performance on WSI classification and segmentation compared with models pre-trained by the natural image dataset (ImageNet \cite{ref25}). 


Despite the successful adoption of SSL, these approaches are not designed to effectively learn image features from multi-resolution WSIs. They often focus on either context or target features separately, thereby disregarding the valuable information offered by multi-resolution WSIs and creating semantic gaps in subsequent stages, wherein global and local information are misaligned. This misalignment during pre-training can adversely affect model convergence in the subsequent fine-tuning stage. To address this problem, we propose a new Dual-branch Self-supervised representation learning Framework for WSI semantic segmentation (DSF-WSI) that simultaneously learns both context and target features. To reinforce the simultaneous learning, we developed a novel \textit{Context-Target Fusion Module (CTFM)} that effectively aligns the learnt multi-resolution features through a \textit{masked jigsaw} pretext task. In this process, target features are randomly masked and shuffled before being concatenated with context features. This formulates a new learning task that enables to extract more sophisticated and hierarchical image features. Additionally, we propose a \textit{Dense SimSiam Learning (DSL)} approach to enhance the extraction of meaningful image features across intermediate layers of the model. We evaluated our framework by comparing it against several State-Of-The-Art (SOTA) supervised and SSL approaches on two public histopathology segmentation datasets. Our proposed DSF-WSI framework has demonstrated superior performance compared to other approaches under the fine-tuning and semi-supervised settings.

\begin{figure}[!t]
\centerline{\includegraphics[width=\columnwidth]{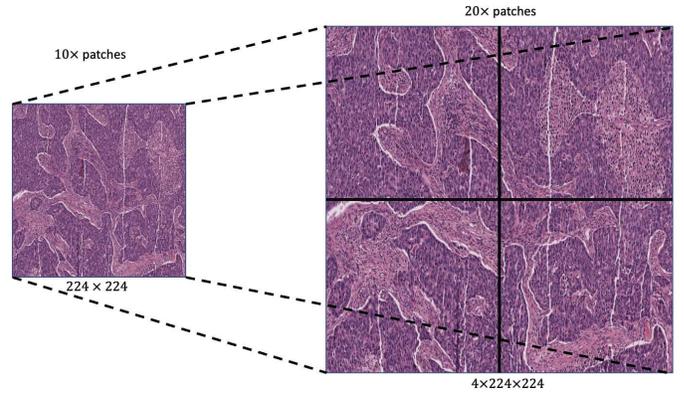}}
\caption{An example of WSI patches extracted from different resolutions. When the sliding window size is fixed at $224 \times 224$, more patches from the high-resolution WSI are required to cover the same field of view as a low-resolution patch. Specifically, in this example, it is necessary to use four $20\times$ patches to achieve the same field of view as one $10\times$ patch.}
\label{fig:ms_imgs}
\end{figure}

\begin{figure*}[t!]
\centerline{\includegraphics[width=\linewidth]{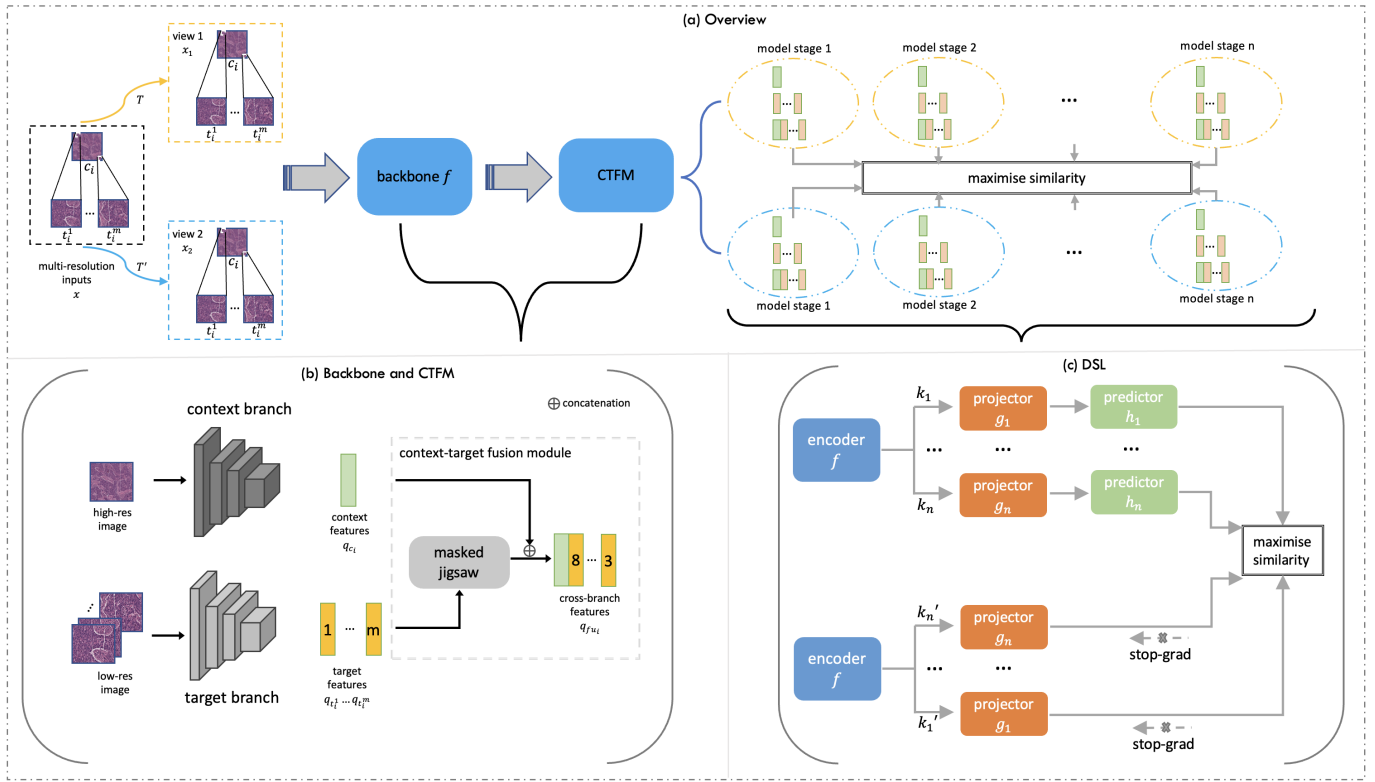}}
\caption{a) Overview of DSF-WSI. Firstly, multi-resolution inputs $x$ are transformed by $T$, $T'$ into two views $x_1$ and $x_2$. Then, they are fed into the backbone network $f$ and refined by the proposed CTFM to generate context features, target features and cross-branch features. After that, DSL is applied to maximise the similarity of these three features produced from each model stage. b) Backbone architecture and CTFM: two same structure networks (do not share weights) are used to process two resolution images separately and generate context features and target features. Then, the CTFM will randomly mask and shuffle target features and concatenate them with context features to derive cross-branch features. c) DSL: for $n$ intermediate model layer features $\{k_1, ..., k_n\}$, they are fed into projectors $\{g_1, ..., g_n\}$ and predictors $\{h_1, ..., h_n\}$ which are then compared with another views $\{k_1', ..., k_n'\}$ to maximise the similarity.}
\label{fig:overview}
\end{figure*}
\section{Related Work}

\subsection{Recent SSL Representation Learning Works}
Self-supervised learning can obtain dataset-specific representations by delving intrinsic data characteristics without the involvement of labels such that fine-tuning these representations could yield better performance with faster convergence. This is usually achieved by defining different pretext tasks, such as context prediction \cite{ref17}, solving jigsaw puzzles \cite{ref18}, image colourisation \cite{ref19} and rotation prediction \cite{ref20}. Recently, the research community has focused more on a variant of SSL, i.e., contrastive learning, which models the similarity and dissimilarity of images \cite{ref5,ref6,ref7,ref22} from different views. Additionally, researchers \cite{ref8,ref23} delved into a more efficient contrastive style which compared the similarity of image views only without measuring their dissimilarity. Despite of promising results reported in these SOTA methods, direct application of them to WSI segmentation can be compromised due to the differences in image statistics, scale, and task-relevant features between natural images and histopathology images.

\subsection{Recent WSI Segmentation Works (non SSL)}
Regardless of different data distributions between natural images and histopathology images, deep models have shown promising results on the WSI segmentation task. For instance, Graham et al. \cite{ref31} proposed a HoVer-Net to generate horizontal and vertical distance maps based on the length of cells to their mass centres. By learning these maps, the model can leverage a shape prior to assist the prediction of segmentation mask. Zhang et al. \cite{ref32} applied a multi-task learning technique by addressing detection and segmentation tasks in parallel. Meanwhile, researchers have also found that using multi-resolution features is beneficial for segmentation performance.
For example, Nir et al. \cite{ref33} extracted image features from different resolutions and integrated them later with support vector machine. Rijthoven et al. \cite{ref13} proposed a multi-branch neural network for processing different resolution images which combined corresponding image feature maps over the branches during the training. Similarly, Schmitz et al.\cite{ref34} developed a family of multi-encoder modules which merged model paths with different WSI resolutions in a spatial relationship-preserving fashion.

\subsection{SSL in WSI Analysis}
One common approach to use SSL in WSI analysis is to simply exchange the ImageNet \cite{ref25} pre-trained extractor with SSL pre-trained models using algorithms such as contrastive predictive coding \cite{ref27}, momentum contrast \cite{ref28} and SimCLR \cite{ref9}. Ciga et al. \cite{ref9} demonstrated the effectiveness of SSL by building a more diverse pre-training dataset that included samples from various histopathology datasets. Similarly, Wang et al. \cite{ref24} designed a hybrid model using Transformer and CNN to extract local-global universal feature representations (i.e., cell-level structures and tissue-level contexts) by pre-training it on a massive dataset containing 15 million unlabelled WSI patches. Moreover, Azizi et al. \cite{ref26} proposed a multi-instance contrastive learning strategy that involved constructing positive pairs using crops from two different images of the same patient case. This approach helped the model learn features that were invariant to both viewpoint and tissue conditions. In contrast, Koohbanani et al. used a multi-task learning approach that involved formulating both domain-agnostic (e.g., image rotation prediction) and domain-specific (e.g., hematoxylin channel prediction) auxiliary tasks. Moreover, Li et al. \cite{ref3} applied SimCLR \cite{ref7} to each of model branch separately and aggregated learnt features later. However, a clear limitation of this approach is the lack of communication among branches during the pre-training process such that context and target branch are learning features independently resulting in a sub-optimal fine-tuning outcome.

\section{Method}

\subsection{Preliminaries: Micros Per Pixel}
The resolution of WSIs is defined by the pixel mapping of the scanned whole slide images, i.e., Microns Per Pixel (MPP). The correspondence between WSI resolution and MPP varies depending on the type of scanners used. In general, $20 \times$ slides refer to images stored at $0.5$ MPP while $40\times$ slides are stored at $0.25$ MPP. Due to these differences of MPP, for different-resolution image patches of the same dimension, low-resolution patches can provide a global architectural composition of the tissue sample, whereas high-resolution patches can offer more specific details about the region of interest and local cellular composition.

\subsection{Overview}
An illustration of our framework is shown in Fig.\ref{fig:overview}. We begin by generating WSI patches at two different resolutions: low and high. These patches are transformed into two distinct views for subsequent SSL tasks: 1) two views of context patches and 2) two views of target patches. These patches are then fed into our dual-branch model to capture resolution-specific image characteristics. We then refine these features further by using our CTFM to establish a path of communication between the branches and obtain cross-branch features. Lastly, the three types of image features (i.e., \textit{context features, target features and cross-branch features}) are then fed into the DSL for WSI tumour representation learning.

\subsection{Contrastive learning method}
\begin{figure}[!t]
\centerline{\includegraphics[width=\columnwidth]{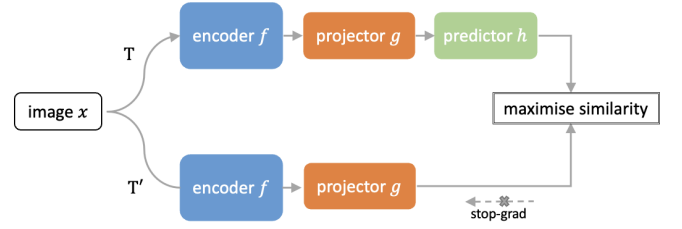}}
\caption{A schematic of SimSiam. The same encoder model $f$ processed two views of an image $x$ generated by two different data transformations, $T$ and $T'$. A projector $g$ and predictor $h$ were then applied to obtain corresponding projection embeddings and output vectors. The objective of SimSiam is to maximise the similarity between $h(g(f(T(x))))$ and $g(f(T'(x)))$.}
\label{fig:simsiam}
\end{figure}

In this paper, we adopt previous SOTA method SimSiam\cite{ref23} as the SSL pre-training strategy. The methodology of SimSiam is shown in Fig.~\ref{fig:simsiam}. Firstly, each image $x$ is randomly transformed via set of augmentations $T$ and $T'$ into two views $x_1$ and $x_2$. These views are then fed into the same encoder model $f$ (i.e., ResNet-18\cite{ref37} in our case) to extract image features. The weights of $f$ are shared between the two views. To minimise the loss of information induced by the contrastive loss function, a multi-layer perceptrons (MLP) projection head $g$ is applied. This is because the objective of the loss function is to make the model invariant to data transformation that may discard useful information for the downstream task, such as colour or orientation of objects. Finally, a prediction MLP head $h$ is adopted after $g$ to transform the output of one view and match it to the other view. We can define the similarity metric as follows:
\begin{equation}\label{eq0}
    \mathcal{D}(p_1, z_2) = - \frac{p_1}{\left\Vert p_1 \right\Vert_2} \cdot \frac{z_2}{\left\Vert z_2 \right\Vert_2}
\end{equation}
where $z_i = g(f(x_i))$ denotes projected embeddings, $p_i = h(z_i)$ denotes predicted output vectors, and $\left\Vert \cdot \right\Vert_2$ is the $l_2$-norm. The symmetric loss for SimSiam can be then defined as:
\begin{equation}\label{eq1}
    \mathcal{L} = \frac{1}{2} \mathcal{D}(p_1, \textit{stopgrad}(z_2)) + \frac{1}{2} \mathcal{D}(p_2, \textit{stopgrad}(z_1)) 
\end{equation}
\textit{stopgrad} is the stop-gradient operation which stops the accumulated gradients from flowing through the operator during the back-propagation. During the back-propagation, the encoder on $x_2$ does not receive gradient updates from $z_2$ in the first term but from $p_2$ in the second term. Vice versa for $x_1$.

\subsection{Multi-resolution SSL pipeline with CTFM}
It is noteworthy that patches extracted from WSIs of different resolutions have varying fields of view. For instance, when a sliding window size of $224 \times 224$ is applied to extract patches from $10 \times$ and $20 \times$ WSI, $20 \times$ patches have only $1/4$ contents of the corresponding $10 \times$ patches, as shown in Fig.~\ref{fig:ms_imgs}. To address this issue, a multi-branch model that processes different image resolutions in separate branches and combines learnt features in later stages can be used.

We followed this direction and designed a dual-branch model consisting of two identical backbone models, with weights not shared. The overview of out SSL pre-training pipeline is depicted in Fig.~\ref{fig:overview}. To learn meaningful WSI representations, three tasks were defined. Firstly, we fed low-resolution patches into the context branch and followed the standard process of SimSiam. Secondly, we processed high-resolution image patches through the target branch, following the same procedure as the context branch. These two tasks helped each branch of the model to extract context and target information contained in the corresponding resolutions respectively. We defined the loss functions as follows:
\begin{equation}\label{eq2}
    \mathcal{L}_{c} = \frac{1}{2} \mathcal{D}(p_c^1, \textit{stopgrad}(z_c^2)) + \frac{1}{2} \mathcal{D}(p_c^2, \textit{stopgrad}(z_c^1)) 
\end{equation}
\begin{equation}\label{eq3}
    \mathcal{L}_{t} = \frac{1}{2} \mathcal{D}(p_t^1, \textit{stopgrad}(z_t^2)) + \frac{1}{2} \mathcal{D}(p_t^2, \textit{stopgrad}(z_t^1)) 
\end{equation}
where $p_c$ and $z_c$ are outputs from context branch, and $p_t$ and $z_t$ are outputs from target branch.

To establish the relationships among branches, we propose the use of cross-branch features, which are generated by associating context and target features. We also introduce a third auxiliary task, called the \textit{masked jigsaw}, which randomly blocks out shuffled image patches. Since a single context image $c_i$ corresponds to multiple target images $\{t_i^1, ... t_i^m\}$, depending on the resolution difference, such as, $1$ of $10 \times$ context patch corresponds to $16$ of $40 \times$ target patches, we can input these images to the backbone to obtain context feature $q_{c_i}$ and target features $\{q_{t_i^1}, ..., q_{t_i^m}\}$. These features were then processed by the proposed CTFM. In this module, target features were shuffled and randomly masked out with a predefined ratio. The cross-branch feature $q_{fu_i}$, which is the output of CTFM, was then derived from the concatenation of context features and processed target features. The objective of our proposed \textit{masked jigsaw} task is to maximise the similarity of cross-branch features from different views as defined by:
\begin{equation} \label{eq4}
    \mathcal{L}_{fu} = \frac{1}{2} \mathcal{D}(p_{fu}^1, \textit{stopgrad}(z_{fu}^2)) + \frac{1}{2} \mathcal{D}(p_{fu}^2, \textit{stopgrad}(z_{fu}^1)) 
\end{equation} 
where $p_{fu}$ and $z_{fu}$ are predicted outputs and projected embeddings derived from $q_{fu}$ respectively. 

By solving these three pretext tasks simultaneously, we enabled the dual-branch model to learn the features of each resolution as well as their interrelationships during the pre-training.

\subsection{Dense SimSiam learning}

The procedure of DSL is illustrated in Fig.~\ref{fig:overview} (c). We define the model stages of encoder ResNet-18 as $S = \{s^1, s^2, s^3, s^4\}$, the corresponding stage features as $K$, and the sets of projectors and predictors as $G$ and $H$, respectively, which output the projected embedding set $Z$ and the predicted vector set $P$. $K, G, Z, H, P$ have the same format as $S$. Thus, we modified Equ.\ref{eq1} and defined the loss function for the $i_{th}$ stage as
\begin{equation} \label{eq5}
    \mathcal{L}^{s^i} = \frac{1}{2} \mathcal{D}(p_1^{s^i}, \textit{stopgrad}(z_2^{s^i})) + \frac{1}{2} \mathcal{D}(p_2^{s^i}, \textit{stopgrad}(z_1^{s^i})) 
\end{equation}
and
\begin{equation} \label{eq6}
    \mathcal{L}_{DSL} = \sum\limits_{i=1}^4 w_i \cdot \mathcal{L}^{s^i}
\end{equation}
where $w$ is the loss weight, and $i$ denotes the stage index.

In summary, there were $12$ projectors and $12$ predictors for the proposed ResNet-18 DSL. And the final loss function was defined as:
\begin{equation} \label{eq7}
    \mathcal{L} = \mathcal{L}_{DSL, c} + \mathcal{L}_{DSL, t} + \mathcal{L}_{DSL, fu}
\end{equation}

\subsection{Fine-tuning and inference}
\begin{figure}[t!]
\centerline{\includegraphics[width=\linewidth]{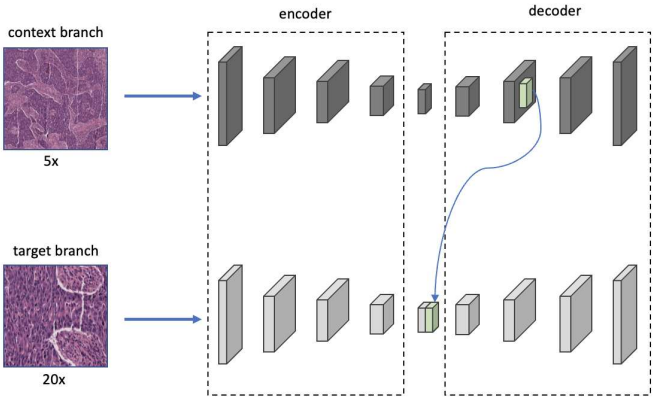}}
\caption{A schematic of HookNet. Skip connections for both branches are omitted for clarity. Feature maps were scaled by a factor of $2$. In this example, the feature maps at depth $2$ in the decoder part of the context branch have the same resolution as the feature maps in the bottleneck of the target branch. Thus, the context and target information were then combined by cropping and concatenation.}
\label{fig:hooknet}
\end{figure}

In this paper, we adopted the previous SOTA work HookNet \cite{ref13} as our baseline method for WSI semantic segmentation task and demonstrated the effectiveness of our algorithm in improving the model segmentation performance. HookNet is a dual-branch encoder-decoder network designed for WSI semantic segmentation using multi-resolution patches. The information from different branches was combined via a "hooking" mechanism, where feature maps in the decoder part from the context branch were cropped and concatenated with the bottleneck feature maps in the target branch, as shown in Fig~\ref{fig:hooknet}. After pre-training the dual-branch backbone by the proposed DSF-WSI, we can simply initialise the encoder part of HookNet and fine-tune it for semantic segmentation task.

\section{Experiments and Discussion}
\subsection{Datasets}
We evaluated our proposed method on two WSI tumour segmentation datasets, the Breast Cancer Semantic Segmentation (BCSS) dataset \cite{ref14} and the Pathology Artificial Intelligence Platform (PAIP) 2019 challenge dataset \cite{ref15}. These datasets were used to estimate model's performance on breast tumour segmentation and liver tumour segmentation, respectively.

\subsubsection{BCSS dataset}
The BCSS dataset \cite{ref14} is a subset of TCGA dataset \cite{ref16}. This dataset consists of 151 hematoxylin and eosin (H\&E) stained WSIs coming from 151 independent breast cancer cases. The annotations were initially collected via crowdsourcing with 25 participants, ranging from senior pathologists to medical students, and were later confirmed by a senior pathologist. The mean size of ROIs is $1.18 mm^2$ ($SD = 0.80 mm^2$) at $0.25$ MPP ($40\times$ magnification). Total of $5$ classes were annotated including Tumour (TUM), Stroma (STR), Lymphocytic inﬁltrate (LYM), Necrosis (NEC) and Other (OTR).

For the data pre-processing, we firstly tiled each WSI using sliding window size of $1024 \times 1024$ with step size of $512$. Based on these tiles, the context patches were generated by directly resizing them into $224 \times 224$ and target patches were generated by using a window size of $256 \times 256$ with a step size of $256$ and then resized into $224 \times 224$. Thus, each context patch ($10\times$ magnification) had $16$ corresponding target patches ($40\times$ magnification). We conducted 5-fold cross-validation (CV) by randomly splitting the dataset with train and validation ratio of $9/1$.

\subsubsection{PAIP2019 dataset} 
The PAIP 2019 dataset \cite{ref15} contains 50 WSIs of liver cancer from 50 patients who underwent resection for hepatocellular carcinoma (HCC) at the Seoul National University Hospital. The slides were stained by H\&E and digitalised with an Aperio AT2 scanner at $20\times$ power and $0.5021 \mu m/px$ resolution, resulting in image sizes between $35,855 \times 39,407$ and $64,768 \times 47,009$ pixels. Two types of annotation are provided: viable regions of cancer cells for continuous tumour areas, as well as whole cancer regions for boundary between the non-tumorous hepatic lobules and the viable tumour (including peritumoral fibrosis, capsules, and inflammation). The initial annotations were provided by a pathologist with 11 years of experience in liver histopathology and reviewed by another expert pathologist. Additionally, we also generated annotations for "tissue area" which indicates healthy tissue pixels by threshold of (R, G, B) $\leq (235, 210, 235)$. This is consistent with the work in \cite{ref34} and allows sampling of healthy tissue patches that can be used as meaningful negative examples.

For the data pre-processing, we generated context ($5\times$ magnification) and target patches ($20\times$ magnification) consistent with the settings used in the BCSS dataset. We randomly selected 10 out of 50 WSIs as the validation set for our CV.


\begin{table}[]
\centering
\caption{A summary of model hyperparameters used in pre-training and fine-tuning on BCSS and PAIP2019.}
\label{tab:tbl0}
\begin{tabular}{@{}lccccc@{}}
\toprule
                                   & \multicolumn{2}{c}{Pre-training}                        &                      & \multicolumn{2}{c}{Fine-tuning}                         \\ \cmidrule(lr){2-3} \cmidrule(l){5-6} 
                                   & \multicolumn{1}{l}{BCSS} & \multicolumn{1}{l}{PAIP2019} & \multicolumn{1}{l}{} & \multicolumn{1}{l}{BCSS} & \multicolumn{1}{l}{PAIP2019} \\ \midrule
\multicolumn{1}{l|}{Epochs}        & 500                      & 300                          &                      & 50                       & 50                           \\
\multicolumn{1}{l|}{Learning rate} & 1e-3                     & 1e-3                         &                      & 1e-3                     & 1e-3                         \\
\multicolumn{1}{l|}{Batch size}    & 32                       & 32                           &                      & 64                       & 64                           \\
\multicolumn{1}{l|}{Optimiser}     & Adam                     & Adam                         &                      & Adam                     & Adam                         \\ \bottomrule
\end{tabular}
\end{table}

\subsection{Model configurations}
We used ResNet-18\cite{ref37} as the backbone of each branch during the pre-training stage. We used the default data transformation settings from SimSiam \cite{ref23} for the contrastive learning phase. The resolution difference between the branches was set to a ratio of $4:1$, and the random masking ratio was set to $0.5$ for our CTFM configuration. As for the DSL, we used a three-layer MLP as the projector, where the hidden dimension and output dimension were set to be equal to the input dimension. Each predictor was a two-layer MLP, and the input and output dimensions were identical, but the hidden dimension was a quarter of the input dimension. The weights of each stage were set to $\{0.1, 0.4, 0.7, 1.0\}$. Other related hyperparameters are shown in Table~\ref{tab:tbl0}. We selected these hyperparameters empirically based on the results of our experiments.

We used the same parameters of HookNet as the original paper \cite{ref13}, where $\lambda$ is set to $1$ to ignore the context loss. Other related hyperparameters are also shown in the Table~\ref{tab:tbl0}.

\begin{table*}[]
\centering
\caption{The results of fine-tuning experiments on BCSS and PAIP2019 using F1 score with 5-fold CV. The best results are in bold.}
\label{tab:tbl1}
\begin{tabular}{@{}lllllllll@{}}
\toprule
\multicolumn{1}{c}{\multirow{2}{*}{Dataset}}   & \multicolumn{1}{c}{\multirow{2}{*}{Method}} & \multicolumn{7}{l}{F1 Score}                                                                                                                                                                                                 \\ \cmidrule(l){3-9} 
\multicolumn{1}{c}{}                           & \multicolumn{1}{c}{}                        & \multicolumn{5}{c}{Cross-validation folds}                                                                                                                                                  &           & Mean (Std)               \\ \midrule
\multicolumn{1}{l|}{\multirow{6}{*}{BCSS}}     & U-Net                                       & 0.7518                              & 0.7647                              & 0.7715                              & 0.7273                              & 0.7306                              &           & 0.7492 (0.0198)          \\
\multicolumn{1}{l|}{}                          & msY-Net                                     & 0.7533                              & 0.7718                              & 0.7755                              & 0.7498                              & 0.7492                              &           & 0.7620 (0.0122)          \\
\multicolumn{1}{l|}{}                          & random-init HookNet                        & 0.7528                              & 0.7472                              & 0.7687                              & 0.7286                              & 0.7316                              &           & 0.7458 (0.0164)          \\
\multicolumn{1}{l|}{}                          & SimSiam-init HookNet                       & 0.7633                              & 0.7876                              & 0.7781                              & 0.7518                              & 0.7594                              &           & 0.7704 (0.0145)          \\
\multicolumn{1}{l|}{}                          & Slf-Hist-init HookNet                     & 0.7876                              & 0.7868                              & 0.7943                              & 0.7585                              & 0.7640                              &           & 0.7782 (0.0159)          \\
\multicolumn{1}{l|}{}                          & DSF-WSI (ours)                              & \textbf{0.8072}                     & \textbf{0.8027}                     & \textbf{0.7991}                     & \textbf{0.7788}                     & \textbf{0.7865}                     & \textbf{} & \textbf{0.7949 (0.0118)} \\ \midrule
\multicolumn{9}{c}{}                                                                                                                                                                                                                                                                                                              \\ \midrule
\multicolumn{1}{l|}{\multirow{6}{*}{PAIP2019}} & U-Net                                       & \multicolumn{1}{c}{0.9168}          & \multicolumn{1}{c}{0.8389}          & \multicolumn{1}{c}{0.9156}          & \multicolumn{1}{c}{0.9039}          & \multicolumn{1}{c}{0.8948}          &           & 0.8940 (0.0321)          \\
\multicolumn{1}{l|}{}                          & msY-Net                                     & \multicolumn{1}{c}{\textbf{0.9276}} & \multicolumn{1}{c}{0.8544}          & \multicolumn{1}{c}{\textbf{0.9335}} & \multicolumn{1}{c}{0.9217}          & \multicolumn{1}{c}{0.9166}          &           & 0.9108 (0.0321)          \\
\multicolumn{1}{l|}{}                          & random-init HookNet                         & \multicolumn{1}{c}{0.9012}          & \multicolumn{1}{c}{0.8992}          & \multicolumn{1}{c}{0.9075}          & \multicolumn{1}{c}{0.8915}          & \multicolumn{1}{c}{0.8927}          &           & 0.8962 (0.0304)          \\
\multicolumn{1}{l|}{}                          & SimSiam-init HookNet                      & \multicolumn{1}{c}{0.9153}          & \multicolumn{1}{c}{0.8983}          & \multicolumn{1}{c}{0.8824}          & \multicolumn{1}{c}{0.9265}          & \multicolumn{1}{c}{0.9261}          &           & 0.9097 (0.0191)          \\
\multicolumn{1}{l|}{}                          & Slf-Hist-init HookNet                     & \multicolumn{1}{c}{0.9245}          & \multicolumn{1}{c}{0.8676}          & \multicolumn{1}{c}{0.9250}          & \multicolumn{1}{c}{0.9193}          & \multicolumn{1}{c}{0.9083}          &           & 0.9089 (0.0241)          \\
\multicolumn{1}{l|}{}                          & DSF-WSI (ours)                             & \multicolumn{1}{c}{0.9260}          & \multicolumn{1}{c}{\textbf{0.9083}} & \multicolumn{1}{c}{0.9304}          & \multicolumn{1}{c}{\textbf{0.9262}} & \multicolumn{1}{c}{\textbf{0.9269}} &           & \textbf{0.9236 (0.0087)} \\ \bottomrule
\end{tabular}
\end{table*}

\begin{table*}[]
\centering
\caption{The results of fine-tuning experiments on BCSS and PAIP2019 using pixel-wise accuracy with 5-fold CV. The best results are in bold.}
\label{tab:tbl2}
\begin{tabular}{@{}lllllllll@{}}
\toprule
\multicolumn{1}{c}{\multirow{2}{*}{Dataset}}   & \multicolumn{1}{c}{\multirow{2}{*}{Method}} & \multicolumn{7}{l}{ACC Score}                                                                                                                                                     \\ \cmidrule(l){3-9} 
\multicolumn{1}{c}{}                           & \multicolumn{1}{c}{}                        & \multicolumn{5}{c}{Cross-validation folds}                                                                                                     &           & Mean \& Std              \\ \midrule
\multicolumn{1}{l|}{\multirow{6}{*}{BCSS}}     & U-Net                                       & 0.9007                     & 0.9059                     & 0.9086                     & 0.8910                     & 0.8923                     &           & 0.8997 (0.0079)          \\
\multicolumn{1}{l|}{}                          & msY-Net                                     & 0.9055                     & 0.9088                     & 0.9102                     & 0.8999                     & 0.8997                     &           & 0.9048 (0.0049)          \\
\multicolumn{1}{l|}{}                          & random-init HookNet                         & 0.9012                     & 0.8992                     & 0.9075                     & 0.8915                     & 0.8927                     &           & 0.8984 (0.0065)          \\
\multicolumn{1}{l|}{}                          & SimSiam-init HookNet                      & 0.9100                     & 0.9151                     & 0.9112                     & 0.9007                     & 0.9038                     &           & 0.9082 (0.0058)          \\
\multicolumn{1}{l|}{}                          & Slf-Hist-init HookNet                     & 0.9151                     & 0.9148                     & 0.9177                     & 0.9034                     & 0.9056                     &           & 0.9113 (0.0064)          \\
\multicolumn{1}{l|}{}                          & DSF-WSI (ours)                             & \textbf{0.9229}            & \textbf{0.9211}            & \textbf{0.9197}            & \textbf{0.9116}            & \textbf{0.9146}            & \textbf{} & \textbf{0.9180 (0.0047)} \\ \midrule
\multicolumn{9}{c}{}                                                                                                                                                                                                                                                                 \\ \midrule
\multicolumn{1}{l|}{\multirow{6}{*}{PAIP2019}} & U-Net                                       & 0.9447                     & 0.8927                     & 0.9440                     & 0.9335                     & 0.9302                     &           & 0.9290 (0.0213)          \\
\multicolumn{1}{l|}{}                          & msY-Net                                     & \textbf{0.9519}            & 0.9031                     & \textbf{0.9559}            & 0.9481                     & 0.9446                     &           & 0.9407 (0.0214)          \\
\multicolumn{1}{l|}{}                          & random-init HookNet                         & 0.9454                     & 0.8956                     & 0.9421                     & 0.9383                     & 0.9339                     &           & 0.9311 (0.0203)          \\
\multicolumn{1}{l|}{}                          & SimSiam-init HookNet                      & \multicolumn{1}{c}{0.9435} & \multicolumn{1}{c}{0.9361} & \multicolumn{1}{c}{0.9218} & \multicolumn{1}{c}{0.9512} & \multicolumn{1}{c}{0.9508} &           & 0.9407 (0.0122)          \\
\multicolumn{1}{l|}{}                          & Slf-Hist-init HookNet                     & 0.9499                     & 0.9119                     & 0.9502                     & 0.9464                     & 0.9391                     &           & 0.9395 (0.0161)          \\
\multicolumn{1}{l|}{}                          & DSF-WSI (ours)                             & 0.9508                     & \textbf{0.9389}            & 0.9538                     & \textbf{0.9510}            & \textbf{0.9514}            &           & \textbf{0.9492 (0.0059)} \\ \bottomrule
\end{tabular}
\end{table*}

\subsection{Evaluation}
We evaluated model performance on the tumour segmentation by F1 score and pixel-wise accuracy score. Suppose we compute the critical values of True Positive (TP), True Negative (TN), False Positive (FP), and False Negative (FN). Then, the F1 score is calculated by 
$$
F_1 = \frac{TP}{TP + \frac{1}{2}(FP+FN)}
$$
and the pixel-wise accuracy score is calculated by
$$
Accuracy = \frac{TP+TN}{TP+TN+FP+FN}
$$

In the BCSS dataset, all $5$ classes were considered including TUM, STR, LYM, NEC and OTR. In the PAIP2019 dataset, all $3$ classes were considered including tissue area, whole tumour area and viable tumour area.

We compared our model with several previous SOTA methods including:
\begin{itemize}
    \item U-Net\cite{ref35}: a single-branch model architecture which is a supervised method for biomedical image segmentation. We used this model directly by feeding target patches only.
    \item msY-Net\cite{ref34}: a dual-branch model architecture which is a recent supervised method for histopathology image segmentation. We used their source code and trained this model from scratch.
    \item HookNet\cite{ref13}: a dual-branch model architecture which is a supervised method for histopathology semantic segmentation. We re-implemented and trained this model from the scratch.
    \item SimSiam\cite{ref23}: a popular SSL algorithm. We applied this algorithm to pre-train two backbones separately on the context patches and target patches, which are then used to initialise the weights of HookNet encodes.
    \item Slf-Hist\cite{ref9}: a recent SSL method proposed for WSI analysis tasks. We used their pre-trained weights to initialise corresponding encoders and compared with our pre-trained model. They pre-trained using a hybrid dataset built with total of 57 datasets consisting of around 4 million patches.
\end{itemize}

Two settings were considered for model performance evaluation:
\begin{itemize}
    \item Fine-tuning setting: models were trained with labels by the full training set and validated by the full validation set.
    \item Semi-supervised setting: models were trained with labels by a fraction (50\%, 10\%and 1\%) of the training set and validated by the full validation set.
\end{itemize}

\subsection{Main results}
\subsubsection{Fine-tuning results}
The results of two datasets with F1 score and accuracy score are shown in Table~\ref{tab:tbl1} and Table~\ref{tab:tbl2}, respectively. For the BCSS dataset, our DSF-WSI achieved the best F1 score of $0.7949$ and the best accuracy score of $0.918$. U-Net\cite{ref35} and random-initialised HookNet\cite{ref13} obtained similar results, with F1 score of $0.7492$ and $0.7458$, respectively. The recent supervised method msY-Net\cite{ref34} outperformed them with F1 score of $0.762$, showing the advantages of the multi-branch architecture design over HookNet. Nevertheless, HookNet initialised with SimSiam-pre-trained weights\cite{ref23} achieved a better performance (F1 score of $0.7704$), demonstrating the effectiveness of SSL-pre-trained weights in histopathology segmentation. Furthermore, by pre-training the model on a large and diverse dataset, the performance was further improved, as shown in the Slf-Hist-initialised \cite{ref9} HookNet, which obtained an F1 score of $0.7782$. Despite of that, our DSF-WSI obtained better results than these different pre-training approaches and demonstrated the importance of learning correlations between branches during the SSL pre-training. The accuracy score had a similar trend to the F1 score but generally obtained higher scores. This is expected since the dataset is imbalanced, with more non-tumour areas than tumours, resulting in higher accuracy scores than F1 scores.

For the PAIP2019 dataset, our method again outperformed other approaches and achieved the best F1 score of $0.9236$ and accuracy score of $0.9492$. Compared with the BCSS dataset, many approaches generally performed better with the PAIP2019 dataset due to larger available training data and easier task settings (i.e., predicting only $3$ classes). One thing to note is that msY-Net\cite{ref34} performed well in this dataset, achieving an F1 score of $0.9108$, which is better than U-Net\cite{ref35} (F1 score of $0.894$), random-initialised HookNet\cite{ref13} (F1 score of $0.8962$), SimSiam-initialised \cite{ref23} HookNet (F1 score of $0.9097$) and Slf-Hist-initialised \cite{ref9} HookNet (F1 score of $0.9089$). Similar to the BCSS dataset, model performance on the accuracy score had a similar trend to the F1 score.

In summary, compared with random-initialised models, our DSF-WSI-initialised model has obtained over $6\%$ F1 score increase on the BCSS dataset and over $3\%$ F1 score increase on the PAIP2019 dataset. Furthermore, our method generated better tumour representations than other SSL methods, i.e., SimSiam \cite{ref23} and Slf-Hist \cite{ref9}, which has shown the effectiveness of CTFM designs and the efficiency of DSL strategy.

\subsubsection{Semi-supervised results}

\begin{table*}[t!]
\centering
\caption{The results of semi-supervised experiments on BCSS and PAIP2019 with 5-fold CV. The best results are in bold.}
\label{tab:tbl3}
\begin{tabular}{llccclccc}
\hline
\multicolumn{1}{c}{\multirow{2}{*}{Dataset}}   & \multicolumn{1}{c}{\multirow{2}{*}{Method}} & \multicolumn{3}{c}{F1 Score}                                          &                    & \multicolumn{3}{c}{Avg ACC Score}                                           \\ \cline{3-9} 
\multicolumn{1}{c}{}                           & \multicolumn{1}{c}{}                        & 50\%                    & 10\%                    & 1\%                     &                    & 50\%                    & 10\%                    & 1\%                     \\ \hline
\multicolumn{1}{l|}{\multirow{6}{*}{BCSS}}     & U-Net                                       & 0.7211(0.0198)          & 0.6327(0.0149)          & 0.5434(0.0430)          &                    & 0.8885(0.0080)          & 0.8531(0.0059)          & 0.8182(0.0163)          \\
\multicolumn{1}{l|}{}                          & msY-Net                                     & 0.7274(0.0191)          & 0.6683(0.0150)          & 0.5559(0.0455)          &                    & 0.8910(0.0076)          & 0.8673(0.0060)          & 0.8224(0.0181)          \\
\multicolumn{1}{l|}{}                          & random-init HookNet                         & 0.7094(0.0199)          & 0.6501(0.0148)          & 0.5363(0.0601)          &                    & 0.8838(0.0080)          & 0.8601(0.0059)          & 0.8152(0.0242)          \\
\multicolumn{1}{l|}{}                          & SimSiam-init HookNet                      & 0.7623(0.0114)          & 0.7326(0.0214)          & 0.6365(0.0312)          &                    & 0.9049(0.0046)          & 0.8931(0.0086)          & 0.8550(0.0119)          \\
\multicolumn{1}{l|}{}                          & Slf-Hist-init HookNet                     & 0.7646(0.0145)          & 0.7465(0.0134)          & 0.6507(0.0244)          &                    & 0.9059(0.0058)          & 0.8986(0.0054)          & 0.8603(0.0098)          \\
\multicolumn{1}{l|}{}                          & DSF-WSI (ours)                             & \textbf{0.7826(0.0121)} & \textbf{0.7572(0.0095)} & \textbf{0.6876(0.0262)} &                    & \textbf{0.9136(0.0039)} & \textbf{0.9031(0.0037)} & \textbf{0.8751(0.0105)} \\ \hline
\multicolumn{9}{c}{}                                                                                                                                                                                                                                                          \\ \hline
\multicolumn{1}{l|}{\multirow{6}{*}{PAIP2019}} & U-Net                                       & 0.8868(0.0309)          & 0.8607(0.0196)          & 0.7868(0.0306)          & \textbf{}          & 0.9247(0.0206)          & 0.9071(0.0131)          & 0.8578(0.0204)          \\
\multicolumn{1}{l|}{}                          & msY-Net                                     & 0.9019(0.0331)          & 0.8598(0.0340)          & 0.8129(0.0191)          & \textbf{}          & 0.9346(0.0221)          & 0.9010(0.0329)          & 0.8753(0.0127)          \\
\multicolumn{1}{l|}{}                          & random-init HookNet                         & 0.8906(0.0312)          & 0.8608(0.0219)          & 0.7970(0.0218)          & \textbf{}          & 0.9272(0.0208)          & 0.9072(0.0146)          & 0.8647(0.0145)          \\
\multicolumn{1}{l|}{}                          & SimSiam-init HookNet                      & 0.9023(0.0281) & 0.8706(0.0435)          & 0.8339(0.0329) & \textit{\textbf{}} & 0.9351(0.0187)          & 0.9139(0.0290)          & 0.8899(0.0221)          \\
\multicolumn{1}{l|}{}                          & Slf-Hist-init HookNet                     & 0.8929(0.0311)          & 0.8765(0.0325)          & 0.8538(0.0139)          & \textbf{}          & 0.9288(0.0207)          & 0.9177(0.0217)          & 0.9027(0.0093)          \\
\multicolumn{1}{l|}{}                          & DSF-WSI (ours)                             & \textbf{0.9150(0.0072)} & \textbf{0.9009(0.0118)} & \textbf{0.8771(0.0098)} & \textit{\textbf{}} & \textbf{0.9436(0.0049)} & \textbf{0.9340(0.0078)} & \textbf{0.9181(0.0065)} \\ \hline
\end{tabular}
\end{table*}

We also evaluated the model performance under the semi-supervised setting, where only partial training data are available during the training. The 5-fold cross-validation results of two datasets with F1 score and accuracy score are shown in Table~\ref{tab:tbl4}. For the BCSS dataset, SSL-pre-trained methods showed advantages over random-initialised models under this setting; however, our method consistently achieved higher performance compared to all other methods. With only $50\%$ labelled data, our DSF-WSI-initialised HookNet obtained an F1 score of $0.7826$ and an accuracy score of $0.9136$, which is superior to other SSL methods with a $0.08$ improvement in F1 score from the random-initialised HookNet. Similarly, under the setting of $10\%$ and $1\%$ labelled data, our method obtained the best F1 scores ($0.7572$ and $0.6876$) and accuracy scores ($0.9031$ and $0.8751$), with greater performance gains when available labelled training data are limited. For instance, the F1 score gap between random-initialised and DSF-WSI-initialised HookNet is $0.1$ for $10\%$ setting and $0.15$ for $1\%$ setting. For the PAIP2019 dataset, our proposed DSF-WSI method continuously outperformed other methods in all semi-supervised settings, achieving F1 scores of $0.915$, $0.9009$, $0.8771$ and accuracy scores of $0.9436$, $0.934$, $0.9181$ respectively.

In summary, our method has demonstrated the robustness of learnt representations, which were efficient for subsequent model fine-tuning using a partial dataset. Specifically, it achieved better results than the fully supervised baseline (i.e., random-initialised HookNet) with $10\%$ of labelled data.

\subsection{Ablation studies}

\subsubsection{Effectiveness of CTFM and DSL}

We evaluated the effectiveness of proposed model components in Table~\ref{tab:tbl4}. All experiments were configured using the same settings as the fine-tuning experiment on the BCSS dataset. We benchmarked the random-initialised model and obtained an F1 score of $0.7471$. We noticed that exclusively using SimSiam-pre-trained weights had mild benefits for the model performance, improving around $0.02$, which was worse than the ImageNet-pre-trained weight (F1 score of $0.7814$). We argue that this underperformance is due to the scale difference of two dataset size (1 million v.s. 7,000), and we require effective designs for the representation learning.

Analogously, our designed CTFM accelerated representation learning and obtaind an F1 score of $0.7881$ ($0.007$ higher than ImageNet-pre-trained). We conjecture that the masked and shuffled target features force the model to exploit related context features to attain the maximum similarity. Additionally, we tested the jigsaw task ($0.7753$) and masking task ($0.7773$) independently and found that combining these two tasks yielded better results.

\begin{table}[h]
\centering
\caption{Ablation experiments on our CTFM and DSL.}
\label{tab:tbl4}
\begin{tabular}{@{}lc@{}}
\toprule
Method                                      & \multicolumn{1}{l}{F1 Score} \\ \midrule
{ImageNet-pre-trained} & {0.7814}      \\ \midrule
Ours (w/o CTFM and DSL, random-init)        & 0.7471                             \\
Ours (w/o CTFM and DSL, simsiam-init)       & 0.7633                             \\
Ours (w/o DSL, only jigsaw)                 & 0.7753                             \\
Ours (w/o DSL, only masking)                & 0.7773                             \\
Ours (w/o DSL, only CTFM)                   & 0.7881                             \\
Ours (w/o CTFM, only DSL)                   & 0.7896                             \\
Ours (with CTFM and DSL)                    & \textbf{0.8072}                    \\ \bottomrule
\end{tabular}
\end{table}

Furthermore, we assessed the effectiveness of the DSL module without CTFM and attained an F1 score of $0.7896$, which shows that enabling early model layers to learn augmentation-invariant representations is beneficial for the SSL performance. We suggest that the standard SimSiam learning only uses features from the last model layer, which usually contains high-level, semantic-relevant representations for the entire input. However, when it comes to histopathology segmentation, the low-level features (e.g., edges, colours and curves) can also be valuable. In addition, since the ROI (i.e., tumour cells) may be present at a small ratio of the image, later model layers cannot learn the ROI features well. Therefore, it is intuitive to enable the feature learning at early stages of the model. 

Lastly, conjoining these two model components gave us around $0.06$ increase from the scratch model and $0.02$ increase from the ImageNet-pre-trained model.

\subsubsection{Robustness of model selection}

We also evaluated the robustness of our proposed algorithm with different model selections in Table~\ref{tab:tbl5}. We benchmarked not only a deeper ResNet-34 \cite{ref37}, but also recent SOTA image recognition models including RegNet \cite{ref46}, EfficientNet \cite{ref47} and SegFormer \cite{ref48} (transformer-based semantic segmentation model). We pre-trained these backbones with our proposed SSL algorithm using the same parameters as the fine-tuning experiment on BCSS dataset, except that the number of epochs was reduced to $300$ to decrease computation time. Afterwards, we substituted them with the encoder part of HookNet.

\begin{table}[h]
\centering
\caption{Ablation study on model backbone.}
\label{tab:tbl5}
\begin{tabular}{lcc}
\hline
Method           & \multicolumn{1}{l}{F1 Score} & \multicolumn{1}{l}{ImageNet Acc} \\ \hline
ResNet-18        & 0.7851                             & 69.758                           \\
ResNet-34        & 0.7880                             & 73.314                           \\
RegNetY-008      & 0.8004                             & 76.314                           \\
EfficientNet-B0  & \textbf{0.8082}                    & \textbf{77.700}                  \\
SegFormer-B0     & 0.7779                             & N/A                              \\ \hline
\end{tabular}
\end{table}

From the Table~\ref{tab:tbl5}, we can observe that the F1 score trend of different backbone models was consistent with their performances on the ImageNet classification task. EfficientNet-B0 obtained the best result of $0.8082$, which was approximately $0.02$ higher than that achieved with ResNet-18 ($0.7851$). It is worth noting that SegFormer-B0 achieved the lowest performance of $0.7779$, even though it was designed for the segmentation task. We suggest that this may be related to the decoupling of the SegFormer decoder, which is optimally designed for the SegFormer encoder part. To enable fair comparisons with previous SOTA methods, we used the standard ResNet-18 as the backbone in our main experiments.

\subsubsection{SSL strategy}
In this paper, we adopted previous SOTA method SimSiam\cite{ref23} as the contrastive learning strategy considering its negative-sample-free property, This property has two important advantages: 1) it disentangles the requirement of large batch size, reducing the GPU memory demand; and 2) it disencumbers the assumption for patch-based WSI methods, where patches from the same WSI should be categorised into the "positive" class. This assumption, however, was ignored in previous works \cite{ref3,ref9,ref10}, causing patches from the same WSI (positive samples) to be mistreated as negative samples if they are allocated into the same mini-batch, hindering the model from receiving correct updates from the calculated loss.

\begin{table}[h]
\centering
\caption{Ablation study on SSL pre-training strategy.}
\label{tab:tbl6}
\begin{tabular}{@{}llllllc@{}}
\toprule
Method  & \multicolumn{5}{l}{} & \multicolumn{1}{l}{F1 Score} \\ \midrule
SimCLR  &    &    &    &   &   & 0.7614                       \\
MoCo v2 &    &    &    &   &   & 0.7642                       \\
SimSiam &    &    &    &   &   & \textbf{0.7671}              \\
BYOL    & \multicolumn{5}{l}{} & 0.7661                       \\ \bottomrule
\end{tabular}
\end{table}

To validate this argument, we evaluated three SOTA SSL algorithms, including SimCLR \cite{ref9}, MoCo \cite{ref6}, and BYOL \cite{ref8}, and report the results in Table~\ref{tab:tbl6}. As expected, negative-sample-free SSL algorithms achieved better results than others, with SimSiam achieving an F1 score of $0.7671$ and BYOL achieving F1 score of $0.7661$.

\subsubsection{Limitations}
There are several limitations to our method that require further consideration for improvement. First, the framework could be generalised to accommodate a wider range of WSI resolutions as inputs and to learn more meaningful information. Secondly, the issue of data imbalance was not addressed during the pre-training stage, and it is crucial to address any biases introduced by this issue. Finally, our proposed SSL pipeline is currently limited to the CNN structure and cannot be directly applied to other network architectures, such as Transformer \cite{ref49}. Future research could address these limitations to further improve the robustness and generalisability of our approach.


\section{Conclusion}

In this paper, we proposed an SSL pre-training framework for WSI tumour segmentation that aims to reduce the burden of data annotation in histopathology. To better exploit the characteristics of multi-resolution WSIs, we developed a dual-branch SSL framework which enables effective connections between branches.Our experimental results on two datasets have shown that our method can effectively extract meaningful WSI segmentation representations and outperform previous SOTA methods. However, there are limitations to our approach that require further improvement, such as accommodating more resolutions of WSIs as inputs, addressing data imbalance issues, and exploring application to other network architectures. We believe that our proposed framework has the potential to significantly reduce the time and effort required for WSI tumour segmentation, and pave the way for more accurate and efficient diagnosis and treatment in the future.

\bibliographystyle{IEEEtran}
\bibliography{refs.bib}

\end{document}